\documentclass[sigconf,nonacm]{acmart}
\usepackage{caption}
\usepackage{multirow}
\usepackage{xcolor}
\usepackage{graphicx}
\usepackage{epstopdf}
\usepackage{url}
\usepackage{comment}
\usepackage{graphicx}
\usepackage{caption}
\usepackage{subcaption}
\usepackage{amsmath} 
\usepackage{tabularx}
\usepackage{enumitem}
\usepackage{multirow}
\usepackage{multicol}
\usepackage{wrapfig}
\usepackage[toc,page]{appendix}

\DeclareMathOperator*{\argmax}{argmax} 
\DeclareMathOperator*{\argmin}{argmin} 

\usepackage{amssymb}

\AtBeginDocument{%
  \providecommand\BibTeX{{%
  \normalfont 
  \kern-0.5em{\scshape i\kern-0.25em b}\kern-0.8em\TeX}
  }}

\begin{document}


\title{Case-based Explainability for Random Forest: Prototypes, Critics, Counter-factuals and Semi-factuals}




\author{Gregory Yampolsky}
\email{gregory.yampolsky@blackrock.com}
\affiliation{%
   \institution{BlackRock, Inc.}
   \city{New York, NY}
   \country{USA}
 }

 \author{Dhruv Desai}
 \email{dhruv.desai1@blackrock.com}
 \affiliation{%
   \institution{BlackRock, Inc.}
   \city{New York, NY}
   \country{USA}
 }

\author{Mingshu Li}
\email{Mingshu.li@blackrock.com}
 \affiliation{%
   \institution{BlackRock, Inc.}
   \city{Atlanta, GA}
   \country{USA}
 }

 \author{Stefano Pasquali}
 \email{stefano.pasquali@blackrock.com}
 \affiliation{%
   \institution{BlackRock, Inc.}
   \city{New York, NY}
   \country{USA}
  }

\author{Dhagash Mehta}
 \email{dhagash.mehta@blackrock.com}
 \affiliation{%
   \institution{BlackRock, Inc.}
   \city{New York, NY}
   \country{USA}
   }

\renewcommand{\shortauthors}{Yampolsky et al.}



\begin{abstract}
The explainability of black-box machine learning algorithms, commonly known as Explainable Artificial Intelligence (XAI), has become crucial for financial and other regulated industrial applications due to regulatory requirements and the need for transparency in business practices. Among the various paradigms of XAI, Explainable Case-Based Reasoning (XCBR) stands out as a pragmatic approach that elucidates the output of a model by referencing actual examples from the data used to train or test the model. Despite its potential, XCBR has been relatively underexplored for many algorithms such as tree-based models until recently. We start by observing that most XCBR methods are defined based on the distance metric learned by the algorithm. By utilizing a recently proposed technique to extract the distance metric learned by Random Forests (RFs), which is both geometry- and accuracy-preserving, we investigate various XCBR methods. These methods amount to identify special points from the training datasets, such as prototypes, critics, counter-factuals, and semi-factuals, to explain the predictions for a given query of the RF. We evaluate these special points using various evaluation metrics to assess their explanatory power and effectiveness.
\end{abstract}
\maketitle

\section{Introduction}
With the rapid deployment of AI applications across various domains in recent years, the field of eXplainable Artificial Intelligence (XAI) has gained significant importance \cite{molnar2020interpretable,linardatos2020explainable}. As AI decisions can have profound consequences, especially in financial and regulated industries, understanding model outputs and decisions is crucial for building trust, ensuring transparency, and facilitating informed decision-making. XAI seeks to unveil the 'blackbox' nature of the AI models, making the processes interpretable through various methods, including feature importance, model-agnostic approaches, and visualization techniques. Among these methods, Explanable Case-Based Reasoning (XCBR) methods have found significant effectiveness or providing better explanations over several other formats\cite{hal_xai,humer2022comparing,kenny2021explaining}. This approach uses key examples to explain prediction for given test data-points, leveraging the human tendency to understand complex concepts through analogies and examples. By doing so, it enhances transparency and compliance, helping stakeholders understand and trust AI decisions. Examples stem from the training samples, and depending on the semantic definition of an example, the information it provides may differ. 

In XAI terminology \cite{sovrano2023objective}, the 'explanandum' is the aspect of the model's decision-making that needs explanation, while 'explanans' (plural 'explanantia') are the tools used to provide this clarity. Prototypes, critics, counter-factuals, and semi-factuals serve as explanantia, aiming either to illuminate specific points (local methods) or to clarify broader dataset distributions (global methods).

Prototypes are straightforward explanantia; they are representative samples that provide an overview of either the dataset or a specific class \cite{bien2011prototype, hart1968condensed}. Approaches such as the k-nearest neighbor (kNN) method retrieve the most similar training samples to aid understanding \cite{zhang2022k,calvo2015improving,kim2014bayesian,li2018deep}, while feature weighting identifies critical features that influence the model's decision-making \cite{gurumoorthy2019efficient,park2004mbnr}, etc. Studies have consistently shown that case-based explanations are more persuasive than alternative approaches \cite{jeyakumar2020can,van2021evaluating}. However, real-world data points are rarely 'clean', and prototypes sometimes fail to represent the data adequately. Critics address these shortcomings by highlighting data points that deviate from typical model predictions, providing insights into areas where prototypes are insufficient \cite{guidotti2022counterfactual}. There are several attempts in the XAI literature to locate critics, for instance, Ref.~\cite{kim2016examples} proposed to employ the Maximum Mean Discrepancy two-sample test for criticism selection.

In parallel to global methods, local methods like counter-factuals and semi-factuals focus on individual decision points, offering insights into how minor modifications could alter model outcomes\cite{hal_xai, holzinger2022explainable, arrieta2020explainable}. Counter-factuals illustrate what might have changed the model’s decision ("if only" scenarios)\cite{keane2021if}, while semi-factuals show how outcomes remain consistent despite potential changes ("even if" scenarios)\cite{aryal2024even}. Intuitively, these methods effectively approximate the decision boundary, enhancing understanding of the model's behavior. Counter-factuals were first viewed as the nearest unlike neighbors based on the natural definition\cite{keane2020good}. Recently, the idea has evolved to include optimization-based\cite{wu2021polyjuice,van2021interpretable}, and heuristic search-driven\cite{schleich2021geco,vermeire2022explainable} methods. Both strategies aim at finding or generating the counter-factuals through minimizing the distance-based cost function accounting for slightly different aspects\cite{guidotti2022counterfactual}. semi-factuals, often considered a special case of counter-factuals, also suggest beneficial recourses. The semi-factual methods can be roughly categorized as\cite{aryal2023even}: feature utility-based approaches\cite{doyle2004explanation}, nearest unlike neighbours\cite{cummins2006kleor}, local-region methods\cite{nugent2009gaining}, and the latest counter-factually guided proposals\cite{aryal2024even}. 

As is seen in the literature, XAI, especially in the realm of case-based explainability, is fundamentally a similarity learning problem. The central task is to identify the most similar or dissimilar cases to a sample under investigation to elucidate the model's output. As discussed in Ref.~\cite{hanawa2020evaluation}, the space in which the distance is computed is of utmost importance. This raises the key question of how to define similarity in a latent space \cite{caruana1999case}. The most intuitive metric, the Euclidean distance or L2 norm, measures the direct line distance between two points. However, Euclidean distance has limitations, particularly in measuring relationships between categorical variables and lacking scale invariance \cite{aggarwal2001surprising}, which are crucial for accurate similarity assessments in diverse datasets. More specifically, if the classifier to be explained is Random Forest (RF), then the RF proximities \cite{breiman-cutler-blog} offer a natural solution to these problems. RFs are well-known for their robust predictive power across varied tasks. More importantly, RF proximities offer local distance measures derived directly from the model, reflecting the data structure and geometry that the model learns—essentially, how the trees diverge based on the features to make predictions. These proximities inherently incorporate feature importance into the similarity measure, thus providing a more accurate representation of the latent feature space. A recent development in this field is presented in Ref.~\cite{proximity_prototypes} which advocates the use of the vanilla RF proximities \cite{breiman-cutler-blog} to select prototypes for each class. This attempt highlighted the great potential of incorporating RF proximities to locate special points within the training samples, however, it was limited to only prototypes and the vanilla RF proximities which is shown not to be an accurate extraction of distance metric from a trained RF.

In the pursuit of refining the capabilities of XCBR for the predictions of RF model, we propose integrating RF proximities into the XCBR framework. This integration aims to identify not only prototypes but also critics, semi-factuals, and counter-factuals within the training samples, ensuring that each identified point maintains coherent plausibility. Tree-based ensembles such as RF have non-smooth and non-differentiable objective functions, posing additional challenges in identifying robust special points \cite{dutta2022robust}. By leveraging the properties of the recently developed RF Geometry and Accuracy Preserving (RF-GAP) proximities \cite{rhodes2023geometry}, this approach enhances the selection process of the explanantia.

The contributions of this paper are threefold, aiming at extending the application and efficacy of XCBR in practical scenarios, especially within the complex data environment like finance: (1) introduce a novel approach for XCBR that utilizes RF-GAP proximities as the primary similarity measure for finding prototypes, critics, semi- and counter-factuals; (2) implement the proposed methodology and rigorously test it on a variety of datasets, including publicly available toy datasets and complex financial applications through a fund classification problem; (3) examine and assess the quality of the identified special points based on various evaluation metrics.

\section{Methodology}
In this Section, we describe the details of definitions and computation for the prototypes, critics, semi- and counter-factuals as well as RF proximities.

\subsection{Case-based Explainability}

Consider a training set $X =\{x_1,x_2,..,x_n\} \subset \mathrm{R}^k$ with labels $y=\{y_1,y_2,...,y_n\} \in \{1,...,L\}$ where $L$ is the number of classes. A local method aims to explain a test query, $q \in X$, while a global method aims to give insight into the entirety of the distribution of $X$.

\subsubsection{Prototypes}
Prototypes are data points that are representative of the distribution of the data \cite{hastie2009elements,hal_xai}. For instance, in the context of loan approvals, prototypes provide representative examples of individuals from each prediction classes: approved and not approved. This allows stakeholders to gain a comprehensive understanding of the entire population by examining just a few key examples.\\
While there is an extensive literature on various methods for identifying prototypes \cite{proximity_prototypes}, our experiments focused exclusively on two approaches: High Density Points (HDP) and a variation of the K-medoids model, referred to as the K-medoids prototypes. Both methods select prototypes based on the training data, $X$, and the distance metric $d$. Prototypes are in the form of a collection of prototype sets $P_l \in X$ for each class $l$, such that $P_1,P_2,..,P_L$ collectively represent $X$ and the corresponding output $y$.\\

\textbf{High Density Points (HDP) prototypes}: The idea of this method is to utilize a proximity matrix which contains proximities between all data pairs. The matrix is partitioned and sorted by proximities. The first prototype of a given class is chosen as the point with the highest sum of proximities to its nearest neighbors of the same class. Once the prototype is located, it is removed from the dataset along with its $k$ nearest neighbors and the process is repeated to find the next prototype.  This process is repeated until the desired number of prototypes is reached.\\

\textbf{K-medoids prototypes}:  Our second method is a deviation of the K-medoids method \cite{k_medoids}. This approach finds prototypes by minimizing the distance between all points to their nearest prototype of the same class; the `medoid` of the cluster now is a prototype.

These methods were favored due to their efficiency in capturing the most statistically significant patterns within data clusters, and higher interpretability by focusing on actual data points as centers\footnote{Moreover, we did not find any publicly available implementation of the methods for RF prototypes proposed in Ref.\cite{proximity_prototypes} to compare our methods with.}

\subsubsection{Critics}
Critics can be thought of as points that are poorly represented by Prototypes. Although the idea of prototypes has been covered extensively in the literature, research on critics is relatively sparse. The primary study we refer is the Maximum Mean Discrepancy (MMD) critic method \cite{kim2016examples,molnar2020interpretable} which utilizes the MMD to find prototypes and critics based on the witness function.
Rather than employing MMD for prototype identification in our approach, we adapted the witness function to derive our critics. The witness function used in our analysis is defined as:
\begin{equation}
    \text{witness}(x)=\frac{1}{n}\sum ^n_{i=1}(prox(x,x_i))-\frac{1}{m}\sum^m_{j=1} (prox(x,z_j)),
\end{equation}
where $n$ is the number of samples in the training set, $m$ is the number of prototypes, $z_j$ is a prototype, and $\text{prox}(x_1,x_2)$ is proximity between $x_1$ and $x_2$. The witness function estimates how much two distributions differ. A higher witness function value indicates that the point is poorly represented by the prototypes, so the critics are points that maximize the witness function, highlighting the area where the model's generalization might not be adequate.

\subsubsection{Semi-factuals}
Given $q$, the semi-factual \cite{aryal2024even, aryal2023even,hal_xai} $x_{sf} \in X$, is defined as:
\begin{equation}
    x_{sf} = \argmax_{x \in X | y(x)=y(q)} d(x,q),
\end{equation}
with $d(x_1,x_2)$ being the distances between $x_1$ and $x_2$.\\
In other words, a semi-factual is the point that is furthest from the query while still maintaining the same label. One can think of a semi-factual as an "even if" explanation. For example, consider someone applying for a loan who wants to understand the decision-making process behind their approval or rejection. A semi-factual explanation would indicate that even if certain aspects were changed, the outcome would remain the same. In this case, it might show that even if the applicant improved their credit score slightly or adjusted their income, they still would not be approved.

\subsubsection{Counter-factuals}
A counter-factual \cite{counterfactual_deficits,keane2021if, hal_xai,mothilal2020explaining}, in contrast to a semi-factual, operates in the opposite direction and is defined as follows: Given a query point, $q$, the counter factual $x_{cf} \in X$, is determined by:
\begin{equation}
    x_{cf} = \argmin_{x \in X | y(x)=y(q)} d(x,q),
\end{equation}
where $d(x_1,x_2)$ represents the distance between $x_1$ and $x_2$.\\
Thus, one can think of a counter-factual as the point closest to the query yielding a different label. This can be conceptualized as an "if only" explanation. Returning to the classic loan refusal example:  if someone was not approved for a loan, a counter-factual explanation might suggest that if only certain conditions were altered—such as increasing the applicant's income or improving their credit score—they would have been approved.

\subsection{Random Forest Proximities}
RF proximities are methods of defining local distances or proximities in  terms of the probabilities of two points sharing terminal leaf nodes.  In this context, the distance between two points can be simply obtained as the complement of their proximity as: $distance = (1-proximity)$. Following the notation in Refs.~\cite{breiman-cutler-blog,rhodes2023geometry}: let $M={(x_1,y_1),(x_2,y_2),..(x_n,y_n)}$ represent the dataset, $T$ being the set of decision trees($t \in T$) in a RF trained on $M$. Call $B(t)$ the representative of the indices of in bag observations for tree $t$, and $O(t)=\{i=1,2,...,n|i\notin B(t)\}$ the representative of the indices of out-of-bag observations for tree $t$. Therefore, $S_i = \{t\in T|i\in O(t)\}$ are the set of trees where the $i^{th}$ sample is out-of-bag (OOB). Let $v_i(t)$ be the set of all indices that end up in the same terminal node as $i$ for tree $t$, and thus $J_i(t)=v_i(t) \cap B(t)$ represent all in-bag samples on the same terminal node. Denote all in bag multiplicities of observation $j$ by $c_j(t)$, the set of all in-bag indices of sharing a terminal node with $i$ with multiplicities by $M_i(t)$, then:
\begin{enumerate}
\item \textit{Original proximity} \cite{breiman-cutler-blog}:
\begin{equation}
    \text{proximity}(x_i,x_j)=\frac{1}{T}\sum_{t=1}^1 I(j\in v_i(t),
\end{equation}
is the probability of $x_i$ and $x_j$ sharing a terminal leaf node.
\item \textit{Out-of-bag proximity} \cite{liaw2002classification}:
\begin{equation}
    \text{proximity oob}(x_i,x_j)=\frac{\sum_{t\in S_i} I(j\in O(t)\cap v_i(t)}{\sum_{t \in S_i} I(j\in O(t)},
\end{equation}
is similar to the prior formula, except this one only involves out-of-bag samples in its calculations.
\item \textit{Geometry and Accuracy Preserving (GAP) proximity} \cite{rhodes2023geometry,lin2006random}:
\begin{equation}
    \text{proximity gap}(x_i,x_j)=\frac{1}{|S_i|}\sum_{t\in S_i}\frac{c_j(t) I(j\in J_i(t)}{|M_i(t)|}.
\end{equation}
Proposed in Ref.~\cite{rhodes2023geometry}, GAP proximity is designed to capture the intrinsic geometry learned by the RF more precisely than previous methodologies. This new definition of proximity appropriately weighs in-bag and out-of-bag samples and thus more closely aligns with the fundamental partitioning that the RF algorithm employs. Notably, GAP proximities provide a closed-form expression that effectively recover RF predictions, offering a robust approach that enhances distance measure across the feature space encoded by the RF (see \cite{jeyapaulraj2022supervised,desai2023quantifying,rosaler2023towards} for their applications in finance). 
\end{enumerate}

\section{Evaluation Metrics}
In this Section, we provide details on the objective evaluation metrics for each of the explanantia \cite{aryal2023even, aryal2024even, molnar2020interpretable, mothilal2020explaining, proximity_prototypes}. 
\subsection{Evaluating Random Forest Training}
For the evaluation of the RF model, we utilized the weighted F1 score for the hold-out test dadtasets. The F1 score is a harmonic mean of precision and recall, providing a balanced measure of the model’s accuracy, especially in the context of imbalanced datasets. 

\subsection{Evaluating Case-based Explainability}
Let $q \in \mathrm{R}^k$ be the explanandum, i.e., the query data-point that needs to be explained, and $e \in \mathrm{R}^k$ be an explanans (either prototype, critics, counter- or semi-factual), where both $q$ and $e$ belong to the training dataset and $k$ is the number of input features in the dataset. Then, the followings are the evaluation metrics used in the present work.

\subsubsection{Distance (semi-factuals and counter-factuals)}
Distance between the explanandum $q$ and the explanans $e$, $d_{e,q}$ is a simple metric that ensures $e$ is close to $q$ for counter-factuals and far from q for semi-factuals.

\subsubsection{Sparsity (semi-factuals and counter-factuals)}
For a counter-factual or semi-factual
to serve as an effective explanans, it is preferable to have minimal feature changes between $q$ and $e$. To quantify this aspect, the sparsity of an explanan is defined as:

\begin{equation}
s_{q,e} = \frac{1}{\text{observed}_{\mbox{f -diff}}}, 
\end{equation}
where $\text{observed}_{\mbox{f-diff}}$ is the number of features changed between the query and explanans. 

It should be noted, however, that this metric is not applicable for proximity nor critics. While it is generally agreed that fewer feature changes is better, there is no concrete evidence that fewer feature changes always lead to better explanations, or that there is an optimal number of features to be changed \cite{keane2021if}.

\subsubsection{Plausibility (all)}
A good explanans should be plausible, i.e., within-distribution. We measure plausibility as the distance between $e$ and the nearest training instance, also known as out-of-distribution (ood) distance: 
\begin{equation*}
\text{ood distance}(e) = \min_{x \in \text{Training Set}} d(e, x).
\end{equation*}
Lower values for ood distance are generally preferred, as they indicate that the explanans $e$ is closer to the distribution of the training data. However, for critics, higher ood distances are better. 
\subsubsection{Confusability(all)}
An explanans (e.g., a semi-factual) should not be confused with another explanans (e.g., a counter-factual), i.e., the classifier should have higher confidence in classifying in the same class of $q$ rather than in a different class (or classes). Confusability of $e$ of a $q$ is measured as: $1 - \text{outlier score(e)}$,
where the outlier score \cite{desai2023quantifying} is calculated as the ratio of the number of in-class samples to the sum of the squared proximity values. Lower values of outlier scores are better with the exception of critics.

\subsubsection{Diversity}
Diversity is a crucial metric when more than one explanantia of the same type (e.g., multiple prototypes) are generated. It is defined as the average pair-wise distance among all the explanantia. Here, a high diversity value indicates a wide range of explanations or coverage of different data regions, ensuring a comprehensive understanding of the dataset through the explanantia.

\subsubsection{Nearest-prototype Test F1 (Only for prototypes)}
A good set of prototypes should be a good descriptor of the class distributions of points in the dataset. In other words, points should belong to the same class as their nearest prototype. Therefore, to evaluate the quality of a set of prototypes, we use a nearest prototype predictor \cite{proximity_prototypes}, which follows similar idea as other nearest neighbor models but focuses on the nearest prototypes. The evaluation process involves using the nearest prototype predictor to classify points in the test set, by assessing the F1 score on the test set, we obtain an unbiased measure of the quality of the chosen prototypes.

\subsubsection{Robustness}
An explanans should be robust to small perturbations of $q$. This robustness can be measured using local Lipschitz continuity \cite{alvarez2018robustness} as:
\begin{equation}
    \text{Robustness}(x) = \argmax_{x_i \in B_e(x)}\frac{||f(X)-f(x_i)||_2}{||x-x_i||_2}.
\end{equation}
A lower value of robustness is better, as it indicates that the explanans is less sensitive to small changes in the input, thus providing more reliable and stable explanations.
\subsubsection{Compactness (only for prototype)}
Compactness of a prototype is defined as the average distance between the prototype and the data-points it represents. A smaller value of compactness indicates that the prototype is a good representative of its corresponding data points, ensuring that the prototype is closely aligned with the data it is meant to represent.

\section{Datasets}

This section covers some of the datasets we will be using.  In order to make sure the results are not biased towards one specific type of data, we compiled results from a variety of datasets.  For identification of prototypes and critics, we used a combination of image-based and tabular datasets. For the generation of semi-factuals and counter-factuals, we focused on tabular datasets. 
\subsection{Toy Datasets}
We have used the following publicly available datasets \cite{asuncion2007uci}.
\begin{itemize}
    \item Diabetes: A small tabular dataset of 393 rows and two classes.  All the columns are numeric.
    \item Blood alcohol: Larger tabular dataset of 2002 rows.  A mixture of numeric and categorical columns.
    \item German credit:  A dataset with many rows that are a combination of numeric and categorical.
    \item MNIST 4\&9: MNIST dataset from \href{https://scikit-learn.org/stable/modules/generated/sklearn.datasets.load_digits.html}{sklearn} filtered only for labels 4 and 9.  Reasoning behind this is that 4 and 9 can easily have overlap and be hard to tell apart.  This is an image based dataset of drawings of the numbers 4 and 9.
\end{itemize}

\subsection{Funds Data} 
We sourced our data for funds from Morningstar Data warehouse data feed. Since Morningstar Categories (the target variable) are based on funds’ portfolio composition, we chose features from the
feed which would describe the same. This dataset provides various levels of aggregation breakdowns which would help explain the funds’ portfolio composition. We chose 'Morningstar Category' as our target class. We then narrowed down the targets to the top 3 categories: 'Large Blend','Large Value','Large Growth'.  This is still a fairly large dataset with 14 numerical and 2 categorical variables, with 4002 funds, for the march 2024 snapshot, and helps us get an idea of how well these methods work on real world financial data for classification. We used one hot encoding for the categorical variables, and interpreted missing values as 0\%. More details about this data can be found in Ref.~\cite{desai2023quantifying}.

\section{Computational Details}
In this Section, we provide computational and implementation details of the computations.
\subsection{Details on Training of Random Forest}
The RF models were optimized specifically for each dataset with 5-fold cross validation using grid search with hyperparameters being number of trees, number of features in each tree and maximum depth.
\subsection{Evaluation Methods}
For each dataset we started with embedding followed by splitting the data into 5-fold cross validation with stratification and balanced class weight for each iteration. The computation for determining example points and evaluation metrics were run on each fold and averaged together afterwards.

\subsubsection{Avoiding biases involving distance types and their respective evaluation metrics}
One issue worth considering is that many evaluation methods rely on some form of distance or proximity. This raises the question: how can we effectively evaluate and compare explanans that are identified using different distances by employing a distance-based metric?  To combat this, we compute the evaluation metrics that depended on distances twice: once dependent on the $L_2$ distance and once dependent on the RF-GAP distance. If an explanans is truly better on an evaluation metric it should perform better in each way that evaluation metric is defined. For example to determine which of the four distance metrics produce the best semi-factual in terms of diversity, we should have that explanans perform better for both $L_2$ diversity and RF-GAP diversity. Otherwise the results can be inconclusive, potentially misleading conclusions about the effectiveness of the distance metrics used.

\subsubsection{Prototypes and critics}
The optimal number of prototypes was determined separately for each method, distance metric, and dataset by conducting a hyperparameter tuning to maximize the f1 score with the prototype count being the hyperparameter. 

After choosing the optimal number of prototypes for each method, dataset, and distance metric, we trained the prototype models to identify the prototypes, and subsequently the matching critics for each configuration using the witness function. 

We then calculated the evaluation metrics for each set of prototypes and critics to assess their performance comprehensively.

\subsubsection{Semi-factuals and Counter-factuals}
Since semi-factuals and counter factuals are inherently local methods they provide a unique output for each query.  To ensure the robustness of our results, we iterated through each point in the dataset as the query, and determined the respective semi- and counter-factuals along with their evaluation metrics. The output includes a collection of five sets —one for each fold of our cross-validation procedure. Each set contained keys associated to lists of factual points and the corresponding evaluation metrics.  To derive the final results, the values were averaged over all points across all folds.

\section{Results}
This section presents the results on the toy datasets and the funds data. The test f1 scores of the RF are summarized in Table \ref{tab:1}.
\begin{table}[t]
\begin{tabular}{c c}
\toprule
   Dataset  & Test F1-score \\ \midrule
    Diabetes &  0.75958 \\
    Blood alcohol & 0.977862\\
    German Credit & 0.668020\\
    MNIST & 0.988912\\
    \bottomrule
\end{tabular}
\caption{F1 scores of the RF models for each dataset}
\label{tab:1}
\vspace{-6mm}
\end{table}

\subsection{Prototypes}
\begin{table*}[t]
\begin{tabular}{c c c c c c c c}
\toprule
    \multicolumn{2}{l}{Dataset} & Diabetes & Blood Alcohol & German Credit & MNIST & Funds & Total\\ 
    Method & Distance type\\ \midrule
     \multirow{4}{*}{k-medoids} &l2&0.638812&0.911608&0.54915&0.983159&0.329989&0.682544\\
&proximity original  &\textbf{0.771115}&0.933144&0.642905&0.985663&\textbf{0.556828}&0.777931\\
&oob proximity&0.74283&\textbf{0.939848}&0.556908&\textbf{0.988884}&0.556589&0.757012\\
&gap proximity  &0.752475&0.929923&\textbf{0.677355}&0.986123&0.554529&\textbf{0.780081}\\
     \midrule
     \multirow{4}{*}{partition} &l2&0.655033&0.854317&0.589765&\textbf{0.98892}&0.38029&0.693665\\
&proximity original&0.758137&\textbf{0.925344}&0.605349&0.986002&\textbf{0.557368}&0.76644\\
&oob proximity original&0.761992&0.863424&0.601498&0.98883&0.556805&0.75451\\
&gap proximity&\textbf{0.784355}&0.90778&\textbf{0.646094}&0.988914&0.555271&\textbf{0.776483}\\
    \bottomrule
\end{tabular}
\caption{F1 scores of prototyeps using each method and distance type}
  \label{tab:2}
\end{table*}

The results presented in Table \ref{tab:2} demonstrate that using GAP proximity yielded the overall best results for prototypes when evaluated based on test F1 score.  It is evident that all the RF proximity-based results consistently outperformed the traditional Euclidean methods, with the exception of the MNIST 4-9 dataset using the partition method.  An interesting observation is that, for some datasets, the nearest prototype predictors even managed to outperform the RF models themselves. This is particularly valuable given that  RF models are computationally expensive compared to nearest prototype predictors.
While the nearest prototype test f1 appears to be the most popular evaluation metric for prototypes, we also assessed various other evaluation metrics as detailed in Table \ref{tab:prototype_toy_data}. The compactness values are all on a very small scale across all four datasets, indicating that the identified prototypes are indeed excellent representations of the corresponding data points. This effectiveness of prototype identification is further validated by comparing the results to those of critics.

\begin{table*}[]
\centering
\begin{tabular}{lllllll}
\toprule
Dataset       & Type       & Method    & Outlier Score                 & Diversity & ood-distance & Compactness \\
\midrule
Diabetes      & Prototypes & k-medoids & 19.46                         & 0.99      & 0.24         & 0.72        \\
              &            & partition & 18.59                         & 0.75      & 0.25         & 0.79        \\
              & Critics   & k-medoids & 93.94                         & 3.94      & 0.51         &             \\
              &            & partition & 92.70                         & 3.87      & 0.51         &             \\ \midrule
Blood Alcohol  & Prototypes & k-medoids & 14.49                         & 1.11      & 0.03         & 0.71        \\
              &            & partition & 13.11                         & 1.45      & 0.04         & 0.78        \\
              & Critics   & k-medoids & 123.85                        & 3.16      & 0.13         &             \\
              &            & partition & 126.09                        & 3.18      & 0.13         &             \\ \midrule
German Credit & Prototypes & k-medoids & 64.10                         & 0.99      & 0.39         & 0.87        \\
              &            & partition & 76.41                         & 2.43      & 0.36         & 0.83        \\
              & Critics   & k-medoids & 316.41                        & 4.09      & 0.58         &             \\
              &            & partition & 322.89                        & 4.08      & 0.60         &             \\ \midrule
MNIST 4\&9     & Prototypes & k-medoids & 4.77                          & 1.00      & 0.07         & 0.33        \\
              &            & partition & 5.03                          & 0.76      & 0.11         & 0.52        \\
              & Critics   & k-medoids & 5.40                          & 3.53      & 0.22         &             \\
              &            & partition & 5.55                          & 3.52      & 0.22         &            \\
              \bottomrule
\end{tabular}
\caption{Results for prototypes and critics for both k-medoids and HDP methods for toy datasets}\label{tab:prototype_toy_data}
\vspace{-6mm}
\end{table*}






\subsection{Criticisms}
The results for critics can be found in Table \ref{tab:prototype_toy_data}. When comparing to prototypes, a clear trend emerges that the outlier scores are significantly higher for critics than for prototypes, particularly in the diabetes, blood alcohol, and German credit datasets. As the outlier score measures the proportion of in-class samples relative to the sum of squared proximity values, these higher outlier scores indicate that the proposed method effectively differentiates between prototypes and critics. A similar trend is observed for out-of-distribution (OOD) distance. The results align with the expectation that OOD distance should be higher for critics, as they are expected to be farther from the distribution of the training data. In terms of diversity, prototypes generally exhibit lower diversity values compared to critics. This observation aligns with their respective roles: lower diversity in prototypes indicates consistency and similarity to the corresponding data, effectively representing the common features within the dataset. Conversely, higher diversity among critics ensure representation of a broad spectrum of unusual cases, providing a thorough view of those edge cases that the model have difficulty categorizing. These findings lead to the conclusion that GAP proximity delivers high-quality results in identifying and distinguishing prototypes and critics across different various, demonstrating its effectiveness and accuracy.

\subsection{Semi-factuals and Counter-factuals}
The entirety of the semi-factuals and counter-factuals results are detailed in Table \ref{tab:semifactual_toy_data}, respectively. When compiling the results across all datasets and all evaluation metrics, several clear patterns are observed. Semi-factuals generally exhibit higher distance values compared to counter-factuals. This aligns with the expectation that semi-factuals are further from the query point $q$ compared to counter-factuals, which are intended to be close to $q$. For sparsity, both semi-factuals and counter-factuals show effectiveness in maintaining or altering predictions with fewer feature changes. This trend is particularly evident in the diabetes and blood alcohol datasets. High sparsity in both types of explanations suggests that minimal changes are required to maintain or flip the model's predictions, supporting their interpretability \cite{keane2020good}. Moreover, both semi-factuals and counter-factuals have higher OOD distances compared to prototypes. This is expected as semi- and counter-factuals approximate the decision boundary of the model, reflecting their greater deviation from the training data distribution. 

Higher OOD distances indicate the magnitude of model’s sensitivity to feature changes and the boundaries where predictions are altered. Furthermore, counter-factuals generally achieve slightly lower robustness, indicating that the RF is less sensitive to feature changes leading to different predictions. An exception is observed in the MNIST dataset, where higher robustness suggest that the model’s predictions change more rapidly with small feature changes \cite{dutta2022robust,artelt2021evaluating}. In addition, both semi-factuals and counter-factuals show high diversity values. For counter-factuals, high diversity is beneficial as it reveals various paths in the feature space leading to different outcomes. For semi-factuals, maintaining adequate diversity ensures that the explanations are not too similar to the original distribution, thus providing meaningful variations. Overall, the examination of semi-factuals and counter-factuals using these evaluation metrics demonstrates the effectiveness and robustness of GAP proximity in identifying significant points. These findings validate the utilization of GAP proximity as a valuable tool for explaining RF predictions.

\begin{table*}[t]
\centering
\begin{tabular}{lllllll}
\toprule
Dataset       & Factual Type    & Distance & Sparsity & ood-distance & Robustness & Diversity \\ \midrule
Diabetes      & Semi-factual    & 0.999449 & 0.126614 & 0.581024     & 59.403326  & 4.018303  \\
              & Counter-factual & 0.614498 & 0.135032 & 0.42876      & 48.828427  & 3.737914  \\
Blood alcohol & Semi-factual    & 0.999995 & 0.218658 & 0.198686     & 4.570338   & 3.396754  \\
              & Counter-factual & 0.825779 & 0.384368 & 0.496959     & 2.678707   & 3.238252  \\
German Credit & Semi-factual    & 0.99946  & 0.068737 & 0.64765      & 5.228149   & 4.128756  \\
              & Counter-factual & 0.684929 & 0.117841 & 0.575762     & 3.944594   & 3.833225  \\
MNSIT         & Semi-factual    & 0.993985 & 0.025582 & 0.507811     & 4.720324   & 3.927984  \\
              & Counter-factual & 0.946577 & 0.026595 & 0.852339     & 15.076175  & 3.575606 \\ \bottomrule
\end{tabular}
\caption{Results for Factuals for toy datasets.}\label{tab:semifactual_toy_data}
\end{table*}



\subsection{MNIST Visualization}
\begin{figure}
\centering
  \includegraphics[width=6cm,height=5cm]{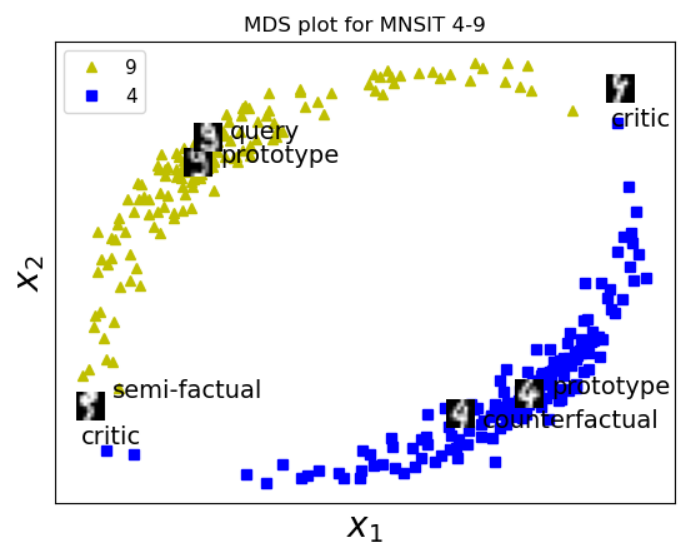}
  \caption{MNIST Visualization (MDS)}\label{fig:MNSIT-vis}
\end{figure}
For visualization we used Multi-Dimensional Scaling (MDS) as our method for dimensionality reduction \cite{borg2007modern}. MDS plots are used to visualize similarity or dissimilarity of data points by mapping them into a lower-dimensional space. The MDS plot in Figure \ref{fig:MNSIT-vis} visualizes the key points in MNIST identified using GAP proximity. The query and all explanans are shown visually. We observe that the optimal number of prototypes is 2, representing the two classes of 4 and 9. The f1 score is high, 0.997, indicating that the prototypes effectively represent the data and suggesting that additional critics may not be necessary for the global explanation. The query point was chosen at random and the semi- and counter-factuals were selected based on the query. Notable, the counter-factual lies on the decision boundary, while the semi-factual is located further away.

\subsection{Funds data}
\begin{figure}
\centering
  \includegraphics[width=9cm,height=5.66cm]{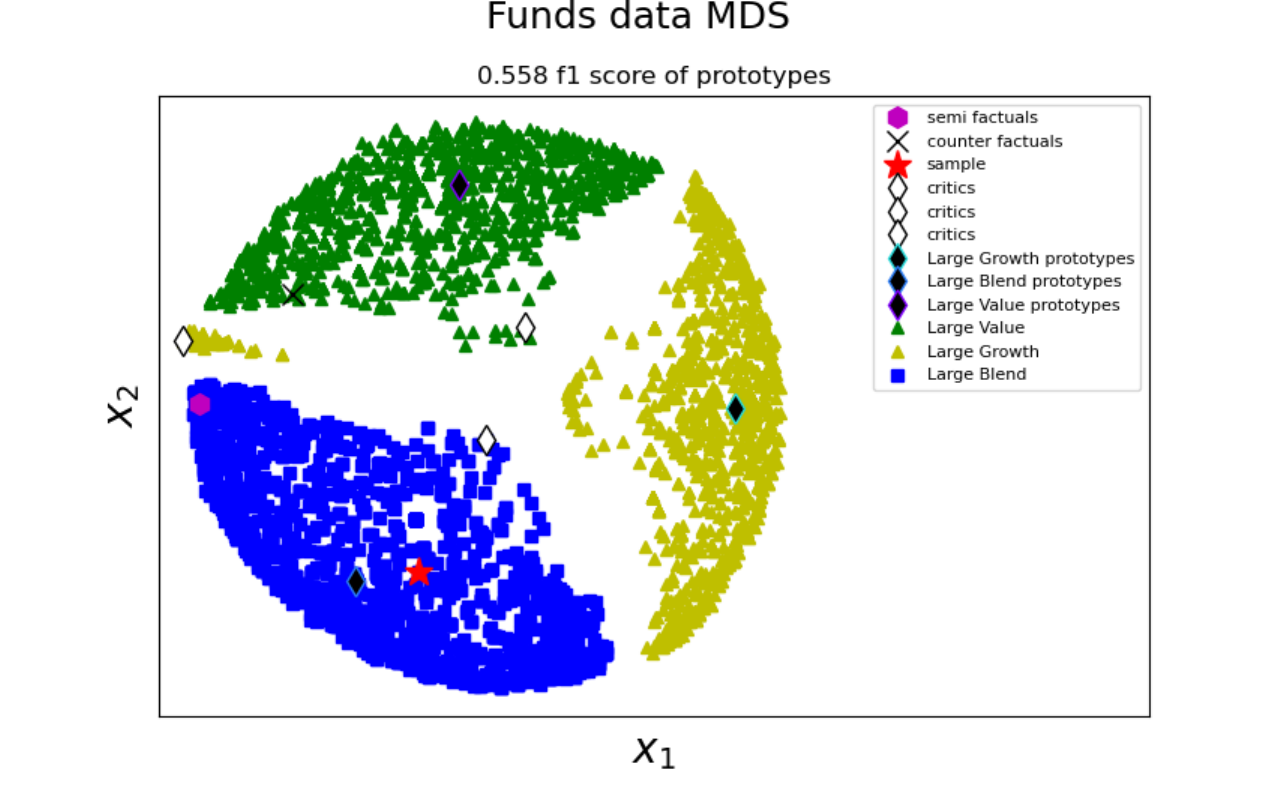}
  \caption{Funds data visualization(MDS)}\label{fig:mds}
\end{figure}
  For the funds data we observed clear groupings using MDS embedding based on the original proximity matrix as the dissimilarity matrix.  The MDS visualization in Figure \ref{fig:mds} clearly shows three prototypes, each located at the center of the three major clusters.  In the Large Growth cluster, there is an outlier group that appears graphically closer to the other two groups than to its own cluster.  However, those outlier points were well captured by the critic.  We randomly selected the query points and identified the resulting semi- and counter-factuals.  Both factual points are meaningful as they explain one of the decision boundaries of the query point.  Overall, the visual support suggests that using RF GAP proximity to find all four of these important points (prototypes, critics, semi- and counter-factuals) is a valid and effective approach.
  \begin{table}[t]
\begin{tabular}{ c c c c c}
\toprule
   Eval metric&Semi&Counter&Prototypes&Critics\\ \midrule
   Distance&0.9998  &0.9908 & na& na\\
    Sparsity &0.014705 &0.018181&na&na\\
    ood distance &0.8874 &0.8795&0.0&0.8361\\
    Robustness &86.34 &89.79 &na&na\\
    outlier score &4944 &373 &40 &5105\\
    diversity &4.537 &0.667 &1.5 &1.4887 \\
    f1 score&na&na&0.558&na\\
    compactness& na& na& 0.7477& na\\
    \bottomrule
\end{tabular}
\caption{Evaluation metrics in terms of RF GAP proximity based distances, for the funds data explanantia}
  \label{tab:3}
  \vspace{-6mm}
\end{table}
  
\section{Conclusion and Outlook}

This research introduced a novel approach to incorporate the RF proximities into the XCBR framework for identifying various types of explanantia including prototypes, critics, semi- and counter-factuals, which play critical roles in explaining RF outcomes.The analysis across various datasets and evaluation metrics has highlighted several key findings. The proposed method successfully identified prototypes and critics, evidenced by metrics such as outlier score, diversity, and out-of-distribution (OOD) distance. Specifically, prototypes, which exhibit lower values in these metrics, effectively represent their data and are central to their corresponding classes. Critics, conversely, display higher values, highlighting atypical cases that challenge the model's decisions. For semi- and counter-factuals, the results confirm the capability of GAP proximity in distinguishing these special points, revealing their roles in approximating the decision boundary, and assessing the model's sensitivity to feature changes.

The findings collectively affirm the great potential of GAP proximity in elucidating the internal mechanisms of RF models, providing comprehensive and effective explanations. The ability of GAP proximity to effectively identify and distinguish between various explanantia, and reveal critical insights into model behavior, makes it a valuable tool for enhancing transparency in AI systems. Future research directions include: (1) refining the application of RF proximities in XCBR to enhance the efficacy of finding more types of explanantia; (2) extending the application to other tree-based ensemble methods to validate its effectiveness and adaptability; (3) investigating the impact of feature space dynamics, such as feature importance, on the searching for explanantia. Such efforts would help in fine-tuning and broadening the method's applicability across various domains beyond finance, leading to more contextually relevant XAI implementations.

\section{Acknowledgement}
The views expressed here are those of the authors alone and not of BlackRock, Inc.

\bibliographystyle{unsrt}

\bibliography{sample-base}


\end{document}